\documentclass{article}
\usepackage{ijcai17}
\usepackage{times}

\usepackage[protrusion=true,spacing=true]{microtype}
\usepackage[group-separator={,}]{siunitx}
\usepackage{amsmath,amsfonts,amsthm}
\usepackage{graphicx,subfig}
\usepackage{enumitem}
\newtheoremstyle{mystyle}
  {} 
  {} 
  {} 
  {} 
  {\bfseries} 
  {.} 
  { } 
  {} 
\theoremstyle{mystyle}

\newtheorem{definition}{Definition}

\newtheorem{theorem}{Theorem}

\newcommand{\tighteq}{\hspace{-3pt}=\hspace{-3pt}}
\newcommand{\tightar}{\hspace{-2pt}\rightarrow\hspace{-2pt}}
\newcommand{\pc}{$\langle \textit{pc} \rangle$}
\newcommand{\modis}{$\langle \textit{modis} \rangle$}
\newcommand{\moral}{$\langle \textit{moral} \rangle$}

\title{Exploiting Causality for Selective Belief Filtering in Dynamic Bayesian Networks (Extended Abstract)\thanks{This paper is an extended abstract of an article in the Journal of Artificial Intelligence Research [Albrecht and Ramamoorthy, 2016].} \vspace{7pt}}

\author{Stefano V. Albrecht \\ The University of Texas at Austin \\ Austin, Texas, United States \\ svalb@cs.utexas.edu \And Subramanian Ramamoorthy \\ The University of Edinburgh \\ Edinburgh, United Kingdom \\ s.ramamoorthy@ed.ac.uk}

\begin{document}

	\maketitle

	\begin{abstract}
Dynamic Bayesian networks (DBNs) are a general model for stochastic processes with partially observed states. Belief filtering in DBNs is the task of inferring the belief state (i.e. the probability distribution over process states) based on incomplete and uncertain observations. In this article, we explore the idea of accelerating the filtering task by automatically exploiting causality in the process. We consider a specific type of causal relation, called \emph{passivity}, which pertains to how state variables cause changes in other variables. We present the \emph{Passivity-based Selective Belief Filtering} (PSBF) method, which maintains a factored belief representation and exploits passivity to perform selective updates over the belief factors. PSBF is evaluated in both synthetic processes and a simulated multi-robot warehouse, where it outperformed alternative filtering methods by exploiting passivity.
	\end{abstract}

	\section{Introduction}

\begin{figure}[t]
	\vspace{-5pt}
	\subfloat[Example DBN]{\label{fig:example-dbn}\includegraphics[height=0.13\textheight]{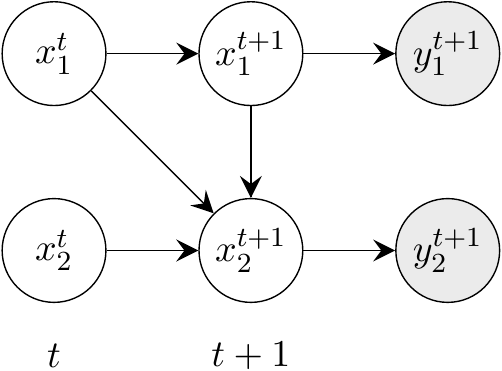}}
	\hspace{17pt}
	\subfloat[Robot arm with 3 joints]{\label{fig:robot-arm}\includegraphics[height=0.135\textheight]{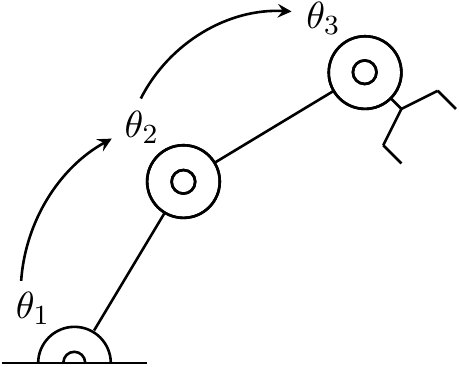}}
	\caption{(a) Variables $x_i^t$ and $x_i^{t+1}$ represent the process state at times $t$ and $t+1$, and $y_j^{t+1}$ represent the observation at time $t+1$. Arrows indicate dependencies between variables.}
\end{figure}

Dynamic Bayesian networks (DBNs) \cite{dk1989} are a general model for stochastic processes with partially observed states (cf. Figure~\ref{fig:example-dbn}). Belief filtering in DBNs is the task of inferring the \emph{belief state}, i.e. the probability distribution over process states, based on incomplete and uncertain observations. This can be a costly operation in processes with large state spaces, requiring efficient approximate methods \cite{kf2009,m2002}.

In this article, we are interested in the application of DBNs as representations of actions in partially observed decision processes, such as POMDPs \cite{klc1998} and their many variants. Decision processes often exhibit high degrees of \emph{causal structure} \cite{p2000}, by which we mean that a change in one part of the process may cause a change in another part. Such causal structure may be used to make the filtering task more tractable, because it can tell us that beliefs need only be revised for certain aspects of the process. For example, if the variable $x_2$ in Figure~\ref{fig:example-dbn} changes its value only if variable $x_1$ changed its value (i.e. a change in $x_1$ \emph{causes} a change in $x_2$), then it seems intuitive to use this causal relation when deciding whether to revise one's belief about $x_2$.

We refer to the above type of causal relation (between $x_1$ and $x_2$) as \emph{passivity}. Intuitively, we say that a state variable $x_i$ is passive in a given action if, when executing that action, there is a subset of state variables that directly affect $x_i$ (i.e. $x_i$'s parents in the DBN representing the action) such that $x_i$ may change its value only if at least one of the variables in this subset changed its value. Passivity occurs naturally in many planning domains, such as in the robot arm shown in Figure~\ref{fig:robot-arm}. If we assume that the joint orientations are absolute (e.g. $\theta_i = 0^\circ$ means that joint $i$ points exactly to the right), then the action of turning joint $i$ leaves variables $\theta_{j \neq i}$ passive, because they change their values only if the corresponding preceding variable $\theta_{j-1}$ changed its value.

How can passivity be exploited to accelerate the filtering task? In the robot arm example, if we choose to rotate joint 3, then the fact that joints 1 and 2 are passive means that they are unaffected by this action. Thus, it seems redundant to revise beliefs for the orientations of joints 1 and 2. However, this is precisely what current filtering methods do. (See the full article for a discussion of related work.)

The purpose of this article is to formalise and evaluate the idea of automatically exploiting causal structure for efficient belief filtering in DBNs, using passivity as a concrete example of a causal relation. We present the \emph{Passivity-based Selective Belief Filtering} (PSBF) method, which maintains a factored belief representation and exploits passivity to perform selective updates over the belief factors. PSBF produces exact belief states under certain assumptions and approximate belief states otherwise. Our method is evaluated in both synthetic processes and a simulated multi-robot warehouse, where it outperformed alternative filtering methods by exploiting passivity.

	\section{Technical Preliminaries}

We consider a decision process which, at each time $t$, is in a state $s^t \in S$ and an agent is choosing an action $a^t$. After executing $a^t$ in $s^t$, the process transitions into state $s^{t+1} \hspace{-1pt} \in S$ with probability $T^{a^t}\hspace{-1pt}(s^t,s^{t+1})$ and the agent receives an observation $o^{t+1} \hspace{-1pt} \in O$ with probability $\Omega^{a^t}\hspace{-1pt}(s^{t+1},o^{t+1})$. We assume factored representations of states and observations, $S = X_1 \times ... \times X_n$ and $O = Y_1 \times ... \times Y_m$, with finite domains $X_i, Y_j$. We write $s_i$ to denote the value of $X_i$ in state $s$, and analogously for $o_j$ and $Y_j$.

The agent chooses action $a^t$ based on its \emph{belief state} $b^t$, which is defined as a probability distribution over the state space $S$ of the process. Belief filtering is the task of updating the belief state $b^t \rightarrow b^{t+1}$ based on the observation $o^{t+1}$.

A \emph{dynamic Bayesian network} \cite{dk1989} for action $a$, denoted $\Delta^a$, is an acyclic directed graph (cf. Figure~\ref{fig:example-dbn}) consisting of:
\begin{itemize}[leftmargin=11pt,itemsep=2pt,label=--]
	\item State variables $X^t \tighteq \left\{ x^t_1,...,x^t_n \right\}$, $X^{t+1} \tighteq \left\{ x^{t+1}_1,...,x^{t+1}_n \right\}$ with $x_i^t,x_i^{t+1} \in X_i$, representing the process states at time $t$ and $t+1$, respectively.
	\item Observation variables $Y^{t+1} \tighteq \left\{ y^{t+1}_1,...,y^{t+1}_m \right\}$ with $y_j^{t+1} \in Y_j$, representing the observation at time $t+1$.
	\item Directed edges $E_a \subseteq \left( X^t \times X^{t+1} \right) \cup \left( X^{t+1} \times X^{t+1} \right) \cup \left( X^{t+1} \times Y^{t+1} \right) \cup \left( Y^{t+1} \times Y^{t+1} \right)$, specifying dependencies between variables.
	\item Conditional probability distributions $P_a(z \, | \, pa_a(z))$ for each variable $z \in X^{t+1} \cup Y^{t+1}$, specifying the probability that $z$ assumes a certain value given a specific assignment to its parents $pa_a(z) = \left\{ z' \, | \, (z',z) \in E_a \right\}$. We also define $pa^t_a(z) = X^t \cap pa_a(z)$ and $pa^{t+1}_a(z) = X^{t+1} \cap pa_a(z)$.
\end{itemize}

\vspace{2pt}

\noindent The functions $T^a$ and $\Omega^a$ are defined via $E_a$ and $P_a$ as
\begin{eqnarray*}
	T^a(s,s') & \tighteq & \prod_{i=1}^n P_a \left( x^{t+1}_i = s'_i \, | \, pa_a(x^{t+1}_i) \hookleftarrow (s,s') \right) \\[5pt]
	\Omega^a(s',o) & \tighteq & \prod_{j=1}^m P_a \left( y^{t+1}_j = o_j \, | \, pa_a(y_j^{t+1}) \hookleftarrow (s',o) \right)
\end{eqnarray*}
where we use the notation $pa_a(x^{t+1}_i) \hookleftarrow (s,s')$ to specify that the parents of $x^{t+1}_i$ in $X^t$ and $X^{t+1}$ assume their corresponding values from $s$ and $s'$, respectively.

	\section{Passivity}

A state variable $x_i^{t+1}$ is called \emph{passive} in action $a$ if there exists a subset of $x_i^{t+1}$'s parents in $X^t$ (in the DBN $\Delta^a$) such that $x_i^{t+1}$ may change its value only if at least one of the variables in this subset changed its value. Formally:

\begin{definition}[Passivity] \label{def:passivity}
	Let action $a$ be specified by DBN $\Delta^a$. A state variable $x_i^{t+1}$ is called \emph{passive} in $\Delta^a$ if there exists a set $\Phi_{a,i} \subseteq pa_a^t(x_i^{t+1}) \setminus \left\{ x^t_i \right\}$ such that:\\[3pt]
	\indent (i) $\forall x_j^t \in \Phi_{a,i} : \left( x_j^{t+1},x_i^{t+1} \right) \in E_a$

	and

	(ii) for any two states $s^t$ and $s^{t+1}$ with $T^a(s^t,s^{t+1}) > 0$\,:
	\begin{equation*}
		\left( \forall x^t_j \in \Phi_{a,i} : s^t_j = s^{t+1}_j \right) \Rightarrow s^t_i = s^{t+1}_i \label{eq:passive}
	\end{equation*}
	
	\noindent A state variable which is not passive is called \emph{active}.
\end{definition}

Clause (i) requires that there is an edge from $x_j^{t+1}$ to $x_i^{t+1}$ for all $x_j^t \in \Phi_{a,i}$. As an example, see Figure~\ref{fig:example-dbn} in which we assumed that the variable $x_2^{t+1}$ was passive with respect to the variable $x_1^t$. Clause (ii) defines the core semantics of passivity by requiring that $x_i^{t+1}$ remain unchanged if all variables in $\Phi_{a,i}$ remain unchanged. Note that this means that the distribution $P_a$ for $x_i^{t+1}$ may specify any deterministic or stochastic behaviour if the variables in $\Phi_{a,i}$ change their values.

\begin{figure}[t]
	\centering
	\includegraphics[height=0.11\textheight]{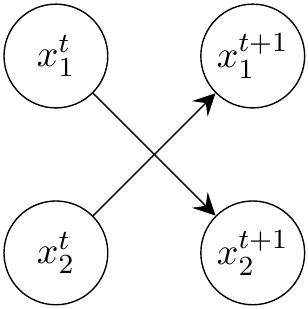}
	\caption{Non-example of passivity.}
	\label{fig:non-example}
\end{figure}

To clarify the role of clause (i), consider the ``non-example'' shown in Figure~\ref{fig:non-example} (observation variables omitted for clarity). The dynamics of the process are such that binary variables $x_1$ and $x_2$ swap their values at each time step. It is easy to verify that both state variables satisfy clause (ii). However, note that these variables are in fact not passive. Passivity is a causal relation and as such it must imply a \emph{causal order} \cite{p2000}. However, there is no causal order between $x_1$ and $x_2$, because there is no edge between $x_1^{t+1}$ and $x_2^{t+1}$. Moreover, passivity means that a variable may change its value only if another variable with respect to which it is passive (a variable in $\Phi_{a,i}$) changed its value. However, the variables in the example depend only on values at time $t$. Clause (i) resolves these issues by requiring that every passive variable $x_i^{t+1}$ must depend on both past and new values of the variables in $\Phi_{a,i}$.

See the full article for a simple procedure which detects passive variables based on their conditional probability tables.

	\section{Passivity-based Selective Belief Filtering}

Passivity-based Selective Belief Filtering (PSBF) uses a two-step update in which the belief state is first propagated through the process dynamics (\emph{transition step} $b^t \tightar \hat{b}^{t+1}$) and then conditioned on the observation (\emph{observation step} $\hat{b}^{t+1} \tightar b^{t+1}$). We focus on the transition step in this abstract since this is where passivity is exploited, and leave the details of the observation step to the full article.

		\subsection{Belief State Representation}

PSBF uses a factored belief state $b(s) \tighteq \prod_{k=1}^K b_k(s_k)$ in which each \emph{belief factor} $b_k$ is a probability distribution defined over the set $S(C_k) = \times_{x_i^{t+1} \in C_k} X_i$ for a cluster $C_k \subseteq X^{t+1}$, such that $C_1 \cup ... \cup C_K = X^{t+1}$. The clustering should be such that strongly correlated variables are in a common cluster while independent or weakly correlated variables are in different clusters \cite{bk1998}. Clusters can be specified manually or generated automatically using methods such as the ones described in Section~6.1 of the full article. See Example~4 in the full article for example clusterings.

		\subsection{Exploiting Passivity in Transition Step}

The idea behind PSBF is to exploit passivity to perform selective updates over the belief factors $b_k$ in the transition step. To do this, we require a procedure which performs the transition step independently for each factor. We obtain such a procedure by introducing two assumptions:
\begin{equation*}
	\begin{array}{ll}
		\text{(A1)} & \forall a : x^{t+1}_i \hspace{-2pt} \in C_k \rightarrow pa_a^{t+1}(x^{t+1}_i) \subseteq C_k \\[5pt]
		\text{(A2)} & \forall k \neq k' : C_k \cap C_{k'} = \emptyset
	\end{array}
\end{equation*}

The first assumption, (A1), states that the clusters must be uncorrelated (i.e. there are no edges in $X^{t+1}$ between clusters), and the second assumption, (A2), states that the clusters must be disjoint. Note that neither assumption implies the other.

Assuming (A1) and (A2), we can perform the transition step $b^t \rightarrow \hat{b}^{t+1}$ independently for each belief factor as
\begin{equation*}\label{eq:exact-new-T}
	\hspace{-55pt} \hat{b}_k^{t+1}(s'_k) = \eta_1 \hspace{-35pt} \sum_{\hspace{36pt} \bar{s} \,\in\, S(pa_{a^t}^t(C_k))} \hspace{-37pt} T^{a^t}_k(\bar{s},s'_k) \hspace{-87pt} \prod_{\hspace{90pt} k' : [ \exists x_i^{t+1} \in C_{k'} \,:\, x_i^t \in \, pa_{a^t}^t(C_k) ]} \hspace{-90pt} b^t_{k'}(\bar{s}_{k'})
\end{equation*}
where $\eta_1$ is a normalisation constant and
\begin{equation*}
	T^a_k(\bar{s},s'_k) = \hspace{-10pt} \prod_{x_i^{t+1} \in\, C_k} \hspace{-8pt} P_a \hspace{-2pt} \left( x_i^{t+1} = (s'_k)_i \ | \ pa_a(x_i^{t+1}) \hookleftarrow (\bar{s},s'_k) \right).
\end{equation*}

The assumptions (A1)/(A2) guarantee that the transition step is performed exactly. They can be violated to obtain approximate belief states, and the full article discusses the various roles of these assumptions.

Given this procedure, we can exploit passivity to perform selective updates over the belief factors $b_k$. Theorem~\ref{th1} provides the formal foundation:
\begin{theorem} \label{th1}
	If (A1) and (A2) hold, and if all $x^{t+1}_i \in C_k$ are passive in $\Delta^{a^t}$, then
	\begin{equation*}
		\forall s \in S : \hat{b}_k^{t+1}(s_k) = b_k^t(s_k).
	\end{equation*}
\end{theorem}

Theorem~\ref{th1} states that if the clusters $C_1,...,C_K$ are disjoint and uncorrelated, and if all variables in cluster $C_k$ are passive in $\Delta^{a^t}$, then the transition step for the corresponding belief factor $b_k^t \rightarrow \hat{b}_k^{t+1}$ can be skipped without loss of information.

How does Theorem~\ref{th1} translate into situations in which (A1)/(A2) are violated? The key assumption is (A1). We can enforce (A1) by modifying the distributions $P_a$ of $x^{t+1}_i \hspace{-2pt} \in C_k$ to marginalise out variables in $pa_{a^t}^{t+1}(x^{t+1}_i)$ which are not in $C_k$, for all clusters $C_k$. However, this modification may cause $x^{t+1}_i$ to lose its passivity property, in the sense that it may no longer satisfy the clauses in Definition~\ref{def:passivity}. Consequently, we would always have to perform the transition step for $C_k$, even if the unmodified variables in $C_k$ are all passive.

To alleviate this effect, one can check if there is a chance that the \emph{unmodified} variables in the cluster change their values. It can be shown that this is the case whenever there is a \emph{causal path} from any active variable to a variable in the cluster:

\begin{definition}[Causal path] \label{def:causalpath}
	A \emph{causal path} in $\Delta^a$, from an active variable $x_i^{t+1}$ to another variable $x_j^{t+1}$, is a sequence $\langle x^{(1)}, x^{(2)}, ..., x^{(Q)} \rangle$ such that $x^{(1)} = x_i^{t+1}, x^{(Q)} = x_j^{t+1}$, and for all $1 \leq q < Q$\,:
	
	\vspace{3pt}
	
	(i) $\left( x^{(q)},x^{(q+1)} \right) \in E_a$

	(ii) $x^{(q+1)}$ is passive in $\Delta^a$ with respect to $x^{(q)}$
\end{definition}

Intuitively, a causal path defines a chain of causal effects (such as between joints 1 and 3 in Figure~\ref{fig:robot-arm}): since the active variable $x^{(q)}$ may have changed its value and $x^{(q+1)}$ is passive with respect to $x^{(q)}$, $x^{(q+1)}$ may also have changed its value, etc. Hence, in the absence of observing these changes, the mere existence of a causal path from $x^{(1)}$ to $x^{(Q)}$ is reason to revise our beliefs about $x^{(Q)}$. Thus, as a general update rule, we can skip the transition step $b_k^t \rightarrow \hat{b}_k^{t+1}$ if all unmodified variables in $C_k$ are passive in $\Delta^{a^t}$, and if there is no causal path from any active variable in $\Delta^{a^t}$ to any variable in $C_k$.

See the full article for a procedure which implements this rule, as well as discussions of computational complexity and approximation errors of PSBF.

	\section{Experimental Evaluation}

The PSBF method was evaluated in synthetic processes with varying sizes and degrees of passivity, as well as a simulation of a multi-robot warehouse system.

		\subsection{Synthetic Processes}

Synthetic processes of four sizes were generated: S (10,3), M (20,6), L (30,9), XL (40,12), where brackets show the number of binary state/observation variables. Each process consisted of two action DBNs which were chosen randomly at each time step. A passivity of $p\%$ means that $p\%$ of state variables were made passive. We used three automatic clustering methods, called \pc, \moral, \modis.

PSBF was compared to a selection of alternative methods: PF \cite{gss1993}; RBPF \cite{dfmr2000}; BK \cite{bk1998}; and FF \cite{mw2001}. For comparison, PF/RBPF/FF were configured to approximate the speed or accuracy of PSBF/BK.

\begin{figure}[t]
	\centering
	\subfloat[0\% passivity]{\includegraphics[height=0.11\textheight]{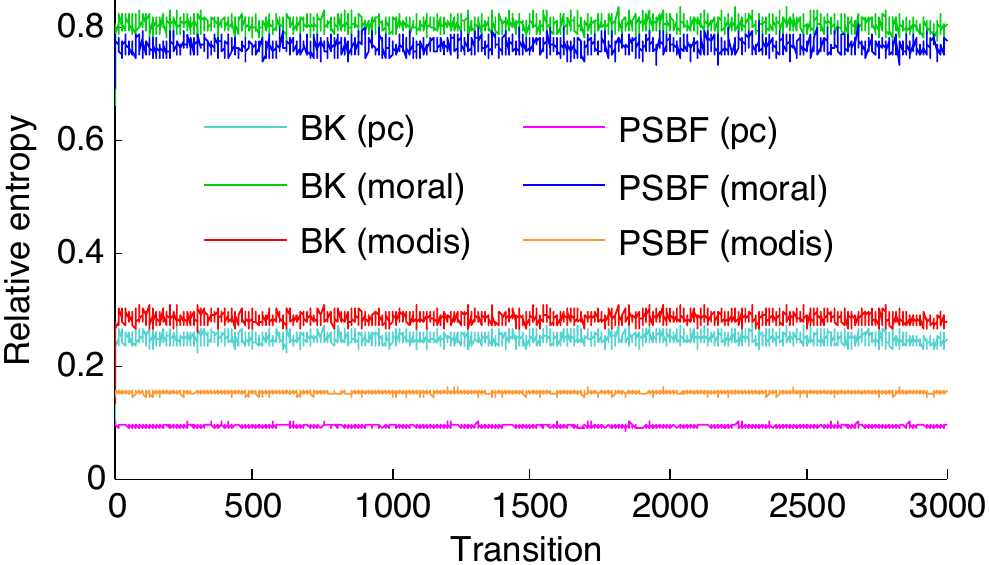}}\hspace{1pt}\subfloat[20\% passivity]{\includegraphics[height=0.11\textheight]{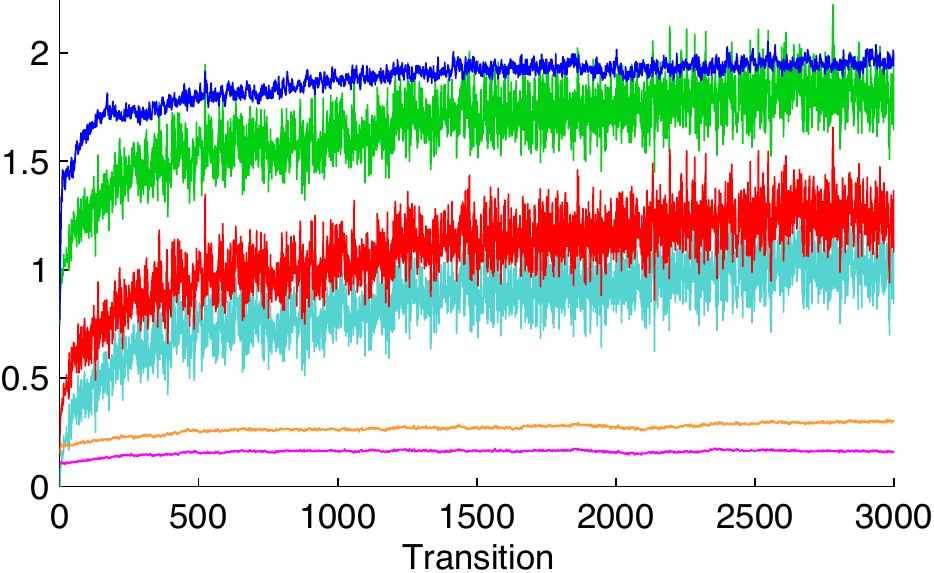}}\\[5pt]
	\subfloat[40\% passivity]{\hspace{4pt}\includegraphics[height=0.11\textheight]{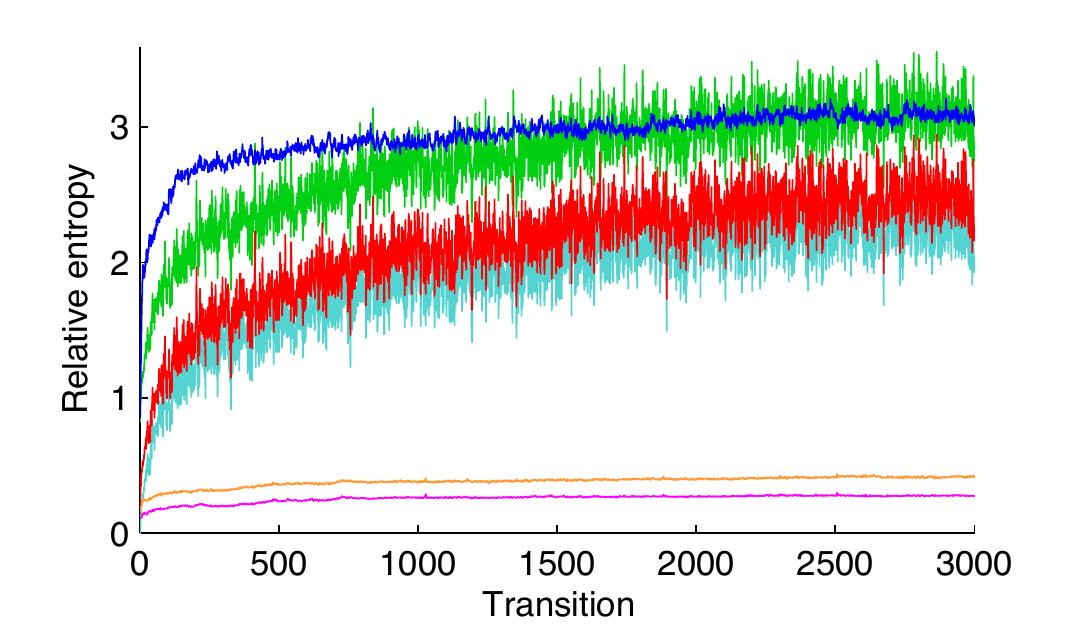}}\hspace{7pt}\subfloat[60\% passivity]{\includegraphics[height=0.11\textheight]{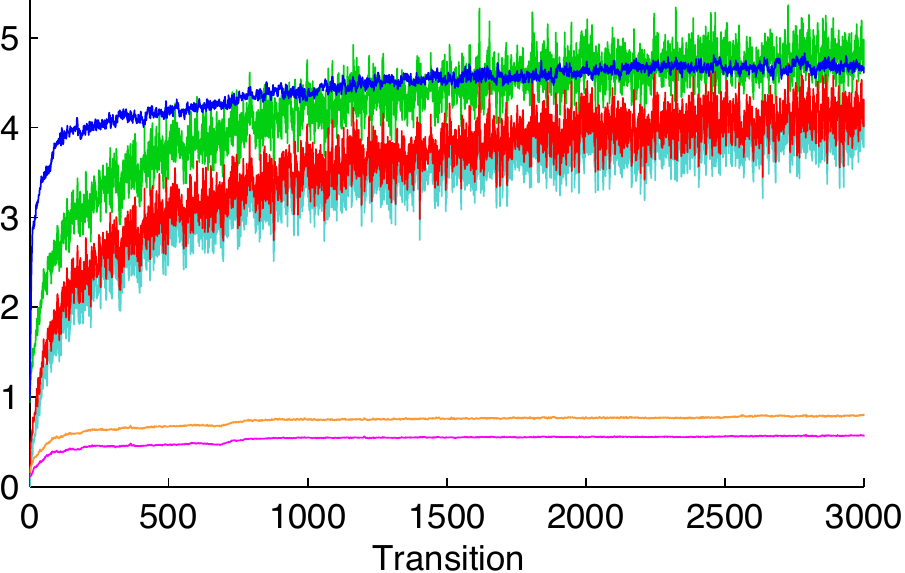}}
	\caption{Accuracy results for synthetic processes of size S ($n=10,m=3$). Relative entropy to exact belief state (lower is more accurate), averaged over 1000 processes.}
	\label{fig:accuracy}
\end{figure}

Figure~\ref{fig:accuracy} shows the relative entropy from exact belief state to PSBF/BK, which achieved the highest accuracy among the tested methods. The results show that PSBF produced a higher or comparable accuracy to BK. They exhibited the same convergent behaviour in relative entropy, showing that the approximation error due to the factorisation was bounded (as discussed in the full article). The relative entropy of both methods increased with the degree of passivity in the process. This is since a higher passivity implies a higher determinacy and, therefore, lower mixing rates, which are a crucial factor in the error bounds of PSBF and BK.

\begin{figure}[t]
	\centering
	\subfloat[S ($n\tighteq10, m\tighteq3$)]{\label{fig:timing-S}\includegraphics[height=0.127\textheight]{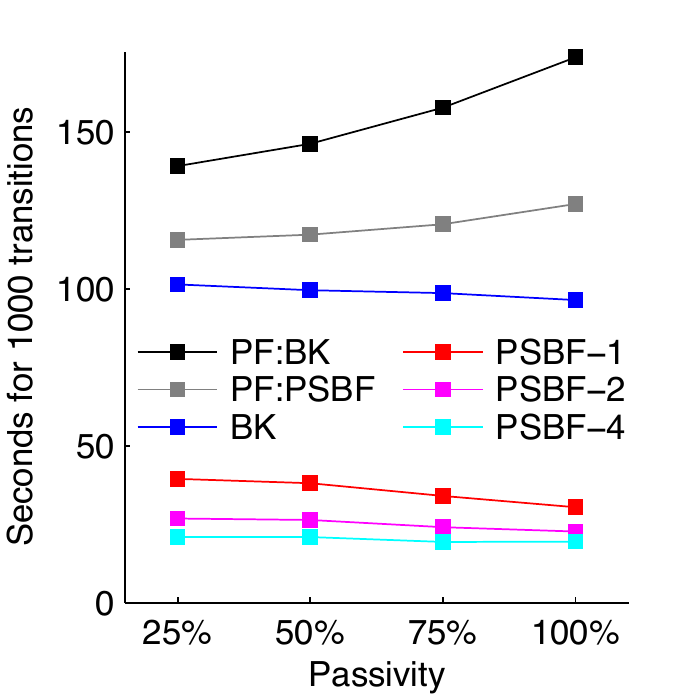}}\hspace{5pt}
	\subfloat[M ($n\tighteq20, m\tighteq6$)]{\label{fig:timing-M}\includegraphics[height=0.127\textheight]{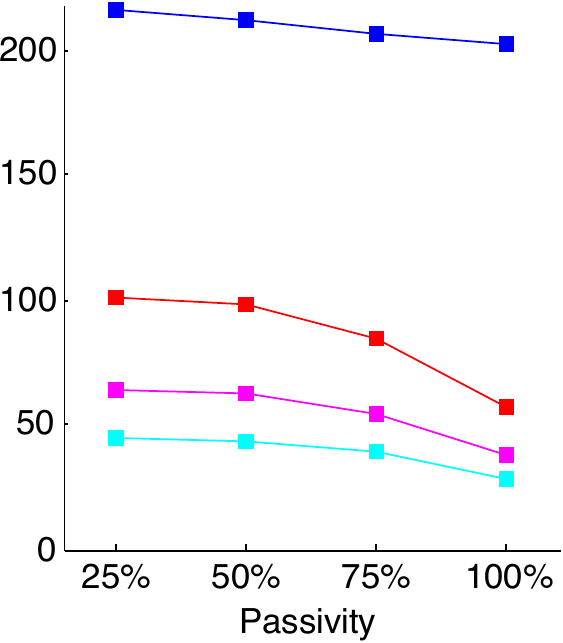}}\hspace{5pt}
	\subfloat[L ($n\tighteq30, m\tighteq9$)]{\label{fig:timing-L}\includegraphics[height=0.127\textheight]{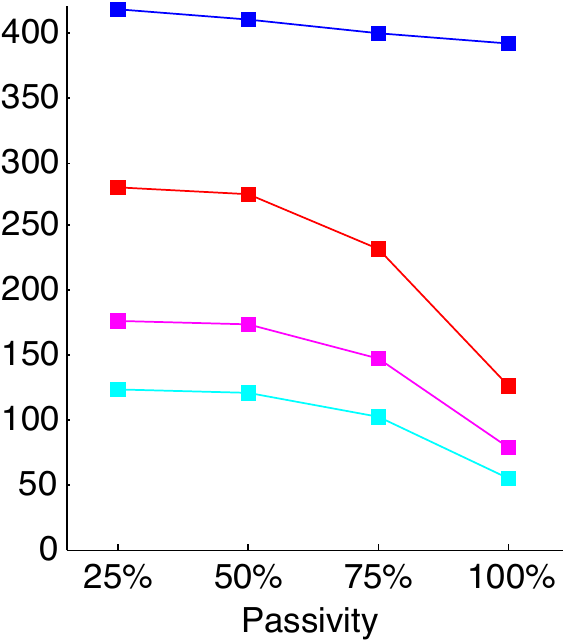}}
	\caption{Timing results for synthetic processes of varying sizes. PSBF run with 1,2,4 parallel threads. PF used number of samples to achieve accuracy of PSBF/BK.}
	\label{fig:timing}
\end{figure}

Figure~\ref{fig:timing} shows computation times (RBPF/FF omitted due to low accuracy results). PSBF was able to minimise the time requirements by exploiting passivity, where the savings grew with both the degree of passivity and the size of the process. While PSBF outperformed BK in our tests, their difference decreased for lower degrees of passivity. With low passivity, PSBF often performed full transition and observation steps, which can be costly operations in large processes. Additionally, the computational overhead of modifying variable distributions and detecting skippable belief factors did not amortise as effectively in large processes with low passivity.

		\subsection{Multi-robot Warehouse}

\begin{figure}[t]
	\centering
	\subfloat[Initial state]{\label{fig:wh-initial}\includegraphics[height=0.13\textheight]{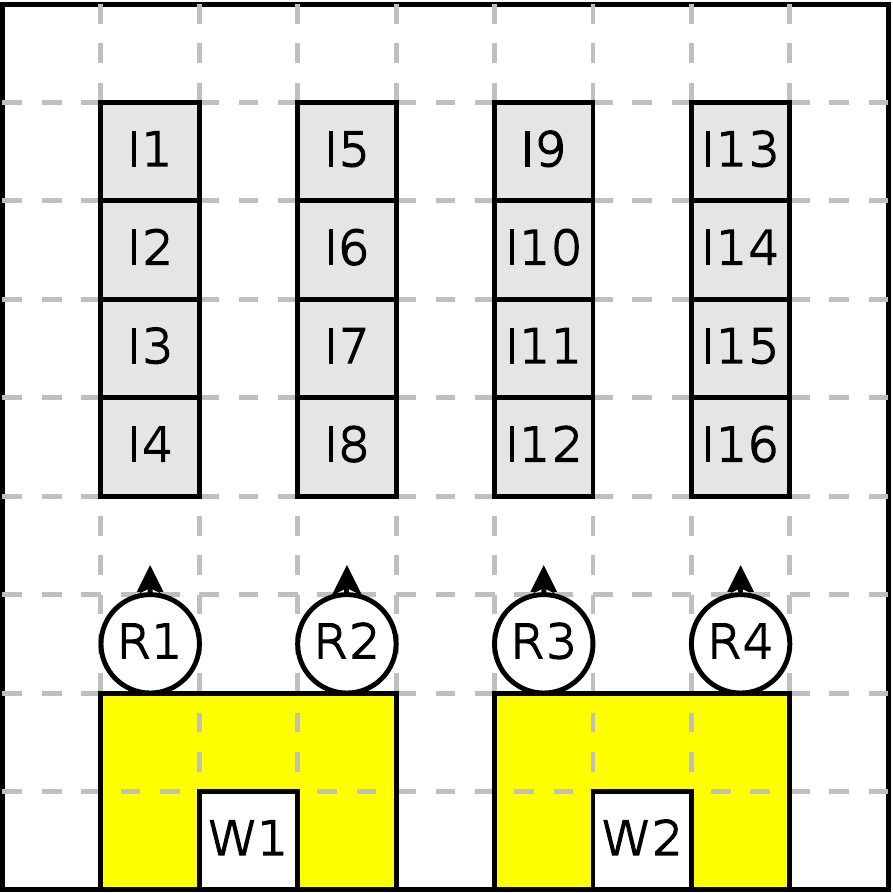}}\hspace{30pt}
	\subfloat[Timing results]{\label{fig:wh-results}\includegraphics[height=0.13\textheight]{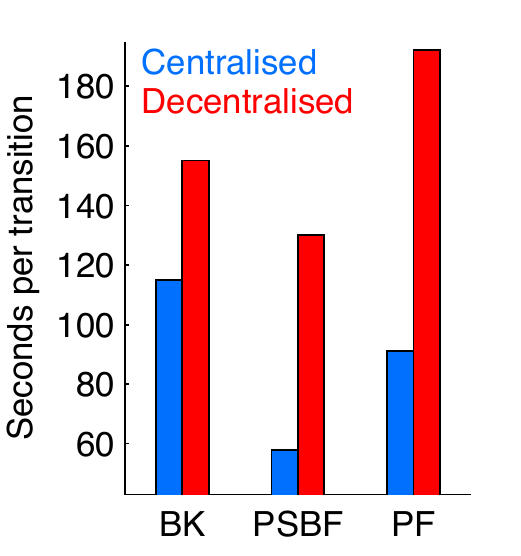}}
	\caption{(a) Multi-robot warehouse consisting of 2 workstations (W1,W2), 4 robots (R1--R4), and 16 inventory pods (I1--I16). (b) Timing for (de)centralised control modes.}
\end{figure}

We demonstrate how passivity can occur naturally in a more complex system, and how PSBF can exploit this to accelerate the filtering task. To this end, we simulated a multi-robot warehouse in the style of Kiva \cite{wdm2008}, in which the robots' task is to transport goods. Figure~\ref{fig:wh-initial} shows the initial state of the simulation. Each robot can perform actions such as moving, turning, and loading pods. Actions and observations have some uncertainty. Each robot maintains a list of tasks such as ``Bring inventory pod I to workstation W'', which are assigned via task auctions. We used two heuristic control modes (centralised and decentralised) to plan actions for robots. See the full article for specifications of DBNs, clusterings, and algorithm configurations.

Figure~\ref{fig:wh-results} shows the time per transition of PSBF, BK, and PF, averaged over 20 different simulations with 100 transitions each. PSBF outperformed BK on average by 49\%/17\% and PF by 36\%/32\% in the centralised/decentralised control mode, respectively. In many cases, PSBF updated less than half of the belief factors, which resulted in significant savings. PSBF's relative savings were smaller in the decentralised mode since its corresponding DBNs had a lower passivity.

The number of states in the warehouse simulation ($\approx 10^{45}$) was too large to compare the accuracy of the tested methods in terms of relative entropy. Instead, we compared their accuracy based on the results of the task auctions and the number of completed tasks in each simulation. In the centralised mode, the algorithms generated over 95\% identical task auctions and completed 15.7 (BK), 15.5 (PSBF), and 15.2 (PF) tasks on average. In the decentralised mode, they generated over 93\% identical auctions and completed 12.1 (BK), 12.2 (PSBF), and 11.7 (PF) tasks on average. These differences were not statistically significant, indicating that PSBF achieved an accuracy similar to that of BK and PF.

	\section{Conclusion}

Our work demonstrates the potential of exploiting causal structure to render the belief filtering task more tractable. In particular, our experiments support the hypothesis that factored beliefs and passivity can be a useful combination in large processes. This insight is relevant for complex processes with high degrees of causality, such as robots used in homes and offices, where the filtering task may constitute a major impediment due to the often very large state space.

There are several directions for future work. For example, it would be useful to know if the definition of passivity could be relaxed while retaining the ability to perform selective updates, and whether the idea of selective inference could be extended to other methods that use factored beliefs. In this work, the selective inference was determined by the parameterisation of the process. An interesting alternative is to frame the selection as a decision problem \cite{as2017aamas}. Ultimately, the key to developing efficient filtering methods is to identify and exploit structure in the process, such as passivity and other recent examples \cite{bg2016,vmb2016}. A grand challenge will be to unite such structural exploitation under one theory of inference.

	\section*{Acknowledgments}

The authors acknowledge the support of the German National Academic Foundation, the UK Engineering and Physical Sciences Research Council (EP/H012338/1), and the European Commission (TOMSY 270436).

	\bibliographystyle{named}
	\bibliography{ijcai17}

\end{document}